\documentclass[letterpaper, 10 pt, conference]{ieeeconf}  % Comment this line out if you need a4paper

\IEEEoverridecommandlockouts  % This command is only needed if 
\usepackage{graphics} % for pdf, bitmapped graphics files
\usepackage{epsfig} % for postscript graphics files
\usepackage{mathptmx} % assumes new font selection scheme installed
\usepackage{times} % assumes new font selection scheme installed
\usepackage{amsmath} % assumes amsmath package installed
\usepackage{amssymb}  % assumes amsmath package installed
\usepackage[bookmarks=true, colorlinks=true, urlcolor=blue]{hyperref}
\usepackage{color}

\definecolor{gr}{rgb}{0, 0.6, 0.1}

\title{\LARGE \bf
Learning Realistic Joint Space Boundaries for Range of Motion Analysis of Healthy and Impaired Human Arms} 

\author{Shafagh Keyvanian$^{1*}$, Michelle J. Johnson$^{1}$, Nadia Figueroa$^{1}$% <-this % stops a space
\thanks{*Corresponding Author}% <-this % stops a space
\thanks{$^{1}$All authors are with the University of Pennsylvania, Philadelphia, PA 19104, USA (e-mail: {\tt\small \{shkey,nadiafig\}@seas.upenn.edu} and {\tt\small johnmic@pennmedicine.upenn.edu})}%
}

\begin{document}
\maketitle
\thispagestyle{empty}
\pagestyle{empty}

%%%%%%%%%%%%%%%%%%%%%%%%%%%%%%%%%%%%%%%%%%%%%%%%%%%%%%%%%%%%%%%%%%%%%%%%%%%%%%%%
\begin{abstract}
A realistic human kinematic model that satisfies anatomical constraints is essential for human-robot interaction, biomechanics and robot-assisted rehabilitation. Modeling realistic joint constraints, however, is challenging as human arm motion is constrained by joint limits, inter- and intra-joint dependencies, self-collisions, individual capabilities and muscular or neurological constraints which are difficult to represent. Hence, physicians and researchers have relied on simple box-constraints, ignoring important anatomical factors. In this paper, we propose a data-driven method to learn realistic anatomically constrained upper-limb range of motion (RoM) boundaries from motion capture data. This is achieved by fitting a one-class support vector machine to a dataset of upper-limb joint space exploration motions with an efficient hyper-parameter tuning scheme. Our approach outperforms similar works focused on valid RoM learning. Further, we propose an impairment index (II) metric that offers a quantitative assessment of capability/impairment when comparing healthy and impaired arms. We validate the metric on healthy subjects physically constrained to emulate hemiplegia and different disability levels as stroke patients. \href{https://sites.google.com/seas.upenn.edu/learning-rom}{[Project webpage]}
\end{abstract}
%%%%%%%%%%%%%%%%%%%%%%%%%%%%%%%%%%%%%%%%%%%%%%%%%%%%%%%%%%%%%%%%%%%%%%%%%%%%%%%%
\section{Introduction}
\label{sec:intro}
Modeling human-like motion is an interesting problem within several research fields, including human-robot interaction (HRI), humanoid control, biomechanics, physics simulations, graphics and computer vision \cite{Akhter2015, Jiang2018, deng2022}. In this work, we focus on human upper-body motions, known to be influenced by constraints due to anatomical and neurological structures, speed vs. accuracy requirements, human capabilities, task constraints, among others \cite{marteniuk87}. 

A critical requirement to achieve a meaningful human upper-body model is a realistic definition of the human joints' range of motion (RoM) considering their inherent anatomical constraints. However, modeling physiological joints constraints is challenging as the RoM of each degree-of-freedom (DoF) depends on multiple anatomical factors; such as (1) minimum and maximum values for each DoF (i.e., box-model boundaries), (2) inter-joint dependencies that define pose-conditioned joint limits (i.e., the RoM of each joint depends on the 3D position of other joints), (3) intra-joint dependencies that define the coupled motion between DoFs in the same joint, and (4) self-collided configurations.
\begin{figure} [tbp]
\begin{minipage}{0.25\linewidth}
\centering
     \includegraphics[width=\linewidth]{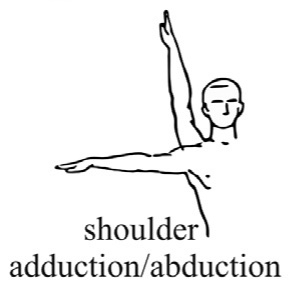}
     \includegraphics[width=0.95\linewidth]{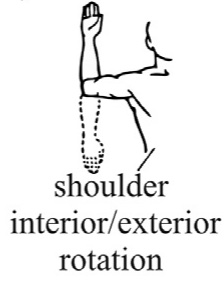}
     \vspace{-10pt}
\end{minipage}\begin{minipage}{0.75\linewidth}
    \centering
     \includegraphics[width=\linewidth]{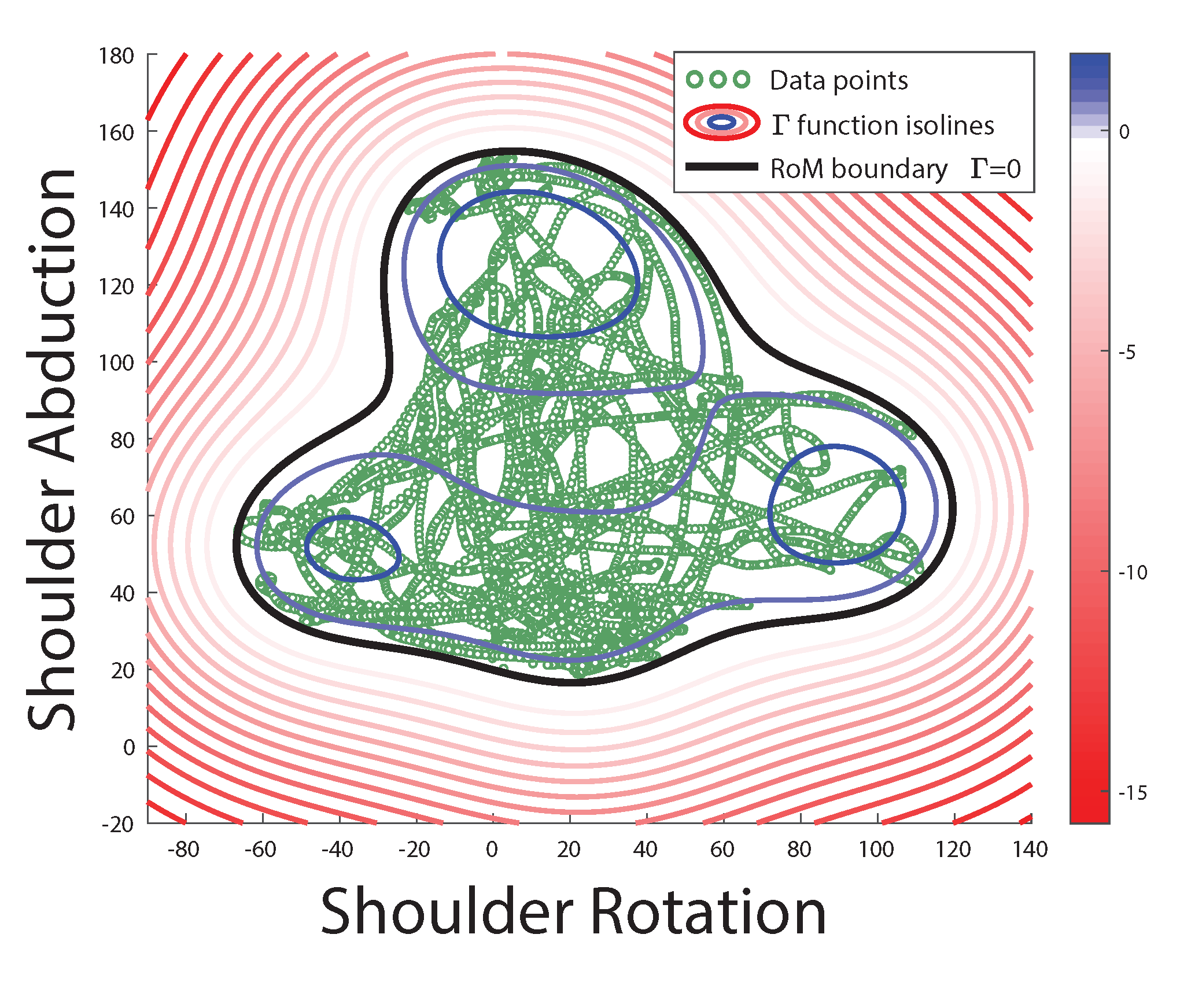}
     % \label{fig:clinical}
\end{minipage}
\vspace{-17pt}
\caption{Realistic (anatomically constrained) RoM boundary for a 2-DoF subset of the human arm (shoulder abduction (deg) vs. rotation (deg)) learned with our proposed joint space RoM boundary $\Gamma(q):\mathbb{R}^N\rightarrow\mathbb{R}$ learning framework. The green dots represent data points collected during experiments (joint-space configuration), The bold black line is the RoM boundary where $\Gamma=0$, and the red ($\Gamma<0$) and blue ($\Gamma>0$) isolines show the regions outside and inside the RoM boundary, respectively.  \label{fig:intro}}
\vspace{-17pt}
\end{figure}
A representation of a human upper-body RoM considering all the aforementioned anatomical factors does not exist. Medical textbooks only offer joint angle limits as box-model boundaries \cite{Hamill2015} for a restricted number of positions \cite{NASA1995}. For instance, in NASA-STD-3000 \cite{NASA1995}, changes in RoM of a few joints are provided when adjacent joints are in 0, 1/3, 1/2, 2/3, and 1/1 ratio of their full limit. We posit that such simplistic and discontinuous joint limits are ill-suited for applications where a realistic RoM is necessary (e.g., HRI applications), since many anatomical factors are disregarded. Further, using generalized joint limits is unrealistic as an individual's capabilities may differ based on demographic features (e.g., age, sex), dominant/non-dominant extremities, life-style, and athletic background. 

For stroke survivors, limited RoM is a common problem, caused by a range of factors (e.g., weakness, spasticity, and joint contractures \cite{Chohan2019}). Joint synergic movement is another common impairment caused by stroke, which occurs when the muscles controlling the joints activate together, leading to stereotyped movement patterns and correlated joints RoM \cite{beer2004, grefkes2011}. Additionally, depending on the severity and location of the stroke, the RoM can vary for each patient. In stroke patients with hemiplegia one side of the body is weakened or paralyzed, leading to different RoM in the impaired and healthy/less-impaired arms. As a result, in safety-critical HRI applications such as robot-assisted rehabilitation, to guarantee safety and to tailor the therapy session to the patient's needs, it is essential to consider the capabilities and impairment level of each patient, hence, modeling the realistic RoM of each patient in a pre-therapy/assessment session is required prior to the robot-assisted therapy session.

Motivated by robot-assisted rehabilitation therapy and impairment diagnosis applications, we propose a data-driven approach to model a realistic RoM from real human data obtained by motion capture. Inspired by prior work on self-collision avoidance boundary learning for multi-DoF robotic systems \cite{mirrazavi2018,koptev2021}, we propose to learn a boundary function, $\Gamma(q):\mathbb{R}^N\rightarrow\mathbb{R}$, to represent a realistic and anatomically constrained RoM for an $N$-DoF human arm defined as:\begin{equation}
\label{eq:gamma-definition}
        \begin{aligned}
            \Gamma(q) > 0 : & ~~\text{Joint Configurations Inside RoM}\\
            \Gamma(q) = 0 : & ~~\text{Joint Configurations on RoM Boundary}\\
            \Gamma(q) < 0 : & ~~\text{Joint Configurations Outside RoM}\\
        \end{aligned}
\end{equation} with $q\in\mathbb{R}^N$ representing the human arm joint configuration. The role of $\Gamma(q)$ is to encapsulate the multi-DoF RoM as a continuous boundary via the level-set $\Gamma(q)=0$, as shown in Fig. \ref{fig:intro}. The $\text{sign}\big(\Gamma(q)\big)$ can be used to i) classify arm configurations as valid (human-like) or invalid in simulation environments and graphics applications (as in \cite{Akhter2015,Jiang2018}) and ii) determine the feasibility of a task for an individual during rehabilitation therapy sessions. Further, as in \cite{mirrazavi2018,koptev2021}, if $\Gamma(q)$ is of class $\mathcal{C}^1$ one can additionally use its gradient $\nabla \Gamma(q)$ to 
iii) enforce the anatomic constraints in solving the inverse kinematics of the human motion, and iv) consider the anatomic constraints within a control policy for robot-assisted rehabilitation (i.e., using the RoM as an invariant set \cite{ames2019}).\footnote{The latter applications are envisioned for future work.} In this work, we showcase the usefulness of our learned RoM approach by additionally proposing a metric that measures the volume of the RoM for each individual for capability/impairment diagnosis. Hence, the \textbf{contributions} in this paper are \textbf{two-fold}: 

\begin{enumerate}
    \item A realistic RoM boundary learning framework that fits a one-class support vector machine (OCSVM) to an anatomically constrained RoM dataset, providing a $\mathcal{C}^1$ boundary function $\Gamma(q)$. 
        
    \item An Impairment Index (II) metric based on the enclosed volume of the learned RoM, validated on a set of experiments designed to represent healthy and impaired arms similar to impairment of stroke patients. 
\end{enumerate}

\noindent \textbf{Paper Organization:} Following we summarize relevant research related to RoM modeling and learning efforts. In Section \ref{sec:learning-rom} we describe our proposed realistic RoM learning framework and introduce the learned RoM Impairment Index metric in Section \ref{sec:impairment-metric}. In Section \ref{sec:experiments} we describe the data collection process, automatic hyper-parameter tuning approach and impairment diagnosis verification. Finally, we discuss future applications for our realistic RoM learned models.

%%%%%%%%%%%%%%%%%%%%%%%%%%%%%%%%%%%%%%%%%%%%%%%%%%%%%%%%%%%%%%%%%%%%%%%%%%%%%%%%

\section{Related Work}
In recent years, upper-limb impairment has been quantitatively assessed by modeling the geometry of the arm's 3D reachable workspace \cite{Lee2020}. While useful to detect functional capabilities, this representation does not offer insight into specific joint level impairments or explicitly considers any of the anatomical joint-space factors mentioned in Section \ref{sec:intro}. Most human model simulation and HRI studies consider human arm joint limits as box-shaped boundaries, including all values within the min/max range of the joints as valid \cite{Akhter2015, Jiang2018}. However, due to the anatomical factors related to joint dependencies and self-collisions, some regions of box-shaped RoM are, in fact, infeasible configurations. This leads to unrealistic human models in physics engines and graphics, and potentially unsafe robot control policies in HRI scenarios. Hence, learning realistic RoM boundaries have become the focus of attention in several research fields.

In computer vision and graphics, learning priors to validate human-like posture has been shown useful to estimate 3D poses from monocular images or depth data. Based on the review done in \cite{Akhter2015, Jiang2018} most methods are insufficient due to the need for additional information or manual intervention, limited to poses in training data, and inaccurate due to unrealistic assumptions. There are also studies that used pose distribution from depth data as priors to generate realistic human poses \cite{grochow2004, min2009, Hauberg2010, zhang2014}. In such distribution-based methods, the boundary of the RoM is not accurately defined, and the constraints are defined based on the probabilities.

Recent studies have tried to address these shortcomings. Akhter et al. \cite{Akhter2015} used motion data to determine the validity of a full-body pose. Since the existing motion capture datasets are insufficient to learn true joint angle limits, the authors in \cite{Akhter2015}, collected more motion data from stretching exercises performed by athletes to include a wider range of human poses. They used the 3D position of joints and, based on the relative child-to-parent joint position in the human skeletal kinematic chain, extracted the local spherical coordinates $(\theta,\phi)$ on a unit sphere. A discretized binary occupancy matrix is then exploited for each bone that yields 1 for valid child-to-parent spherical angles while, in the case of invalid bone segments, all the offspring bones are considered invalid. Akhter's method considers inter-joint dependencies and provides accurate human-like pose detection due to its more inclusive dataset. However, the derived look-up-table function is i) discontinuous, making it difficult to use within gradient-based motion optimization problems, and ii) the joint constraints are not interpretable to physicians that are used to biomechanics definition, as they are spherical angles.

Building upon Akhter’s work, Jiang et al. \cite{Jiang2018} proposed an algorithm that learns an analytical and differentiable model that can be incorporated in problems that demand boundary gradients. In \cite{Jiang2018}, an implicit equation is proposed to represent the boundary of valid human joint configurations. The method produces configurations from random shoulder and elbow angles, solves the joint positions via forward kinematics, and labels each configuration using Akhter's look-up validity function (prior). The generated data is then used to learn a boundary function using a fully connected neural network (NN). While providing a differentiable boundary function, this method suffers from Gimbal lock of Euler angles, and also non-contiguous clustering of valid poses in joint space. This leads to mis-classifying invalid joint configurations that result as valid joint positions. To resolve this ambiguity, they enforced standard box constraints on the random input angles (i.e. $\left[ - 60^\circ,120^\circ \right]$ for shoulder rotation and $\left[ 0^\circ,180^\circ \right]$ for elbow), which limits the actual values in human RoM. 
In both approaches, the inter-joint dependency of siblings in the kinematic chain and the intra-joint coupling for each joint are ignored. Moreover, the twist around the longitudinal axis is also ignored as the classification algorithm is implemented on spherical coordinates calculated from Cartesian coordinates. This neglect is an issue, as intra-joint dependencies rely on the twist component of the rotation.

There are several well-known rotation parametrizations which have advantages and drawbacks with respect to the intended application (e.g., Euler angles, unit quaternion, axis-angle vector, and twist-swing) \cite{Baerlocher2001}. The axis-angle and unit quaternions cannot reflect an intuitive decomposition of the angular motion components. In Euler angles representation the rotation sequence can be defined so that the third angle is used to perform the twist motion. However, as mentioned earlier it leads to a dis-contiguous cluster of valid poses in joint space and often causes the problem of Gimbal lock. 

Herda et al. \cite{Herda2005} determined the shoulder and elbow RoM considering inter-joint and shoulder's intra-joint dependencies. The authors used the quaternion field representation of joint orientations and derived a continuous implicit surface approximation for the quaternions. Their quaternion representation is not subject to singularities such as the Gimbal lock of Euler angles, or mapping rotations of 2$\pi$ to zero rotation. However, the scalar part (cos $\theta$) of the quaternion is deduced, leading to ambiguity in rotation angles. Quaternions also have dual representation ($+q$ and $- q$) for a given joint rotation, and pre-processing on the data over motion trajectory is required to resolve this issue \cite{Murthy2019}. 

In twist-swing parametrization, the first two Euler angles are replaced by an axis-angle vector with zero component along the longitudinal axis which resolves the singularities of Euler angles \cite{Baerlocher2001,Dobrowolski2015}. In Murthy et al.~\cite{Murthy2019} the twist-swing representation in exponential maps form is used to derive the full-body joint limits considering both inter- and intra-joint dependencies. A set of NN discriminators is trained with synthetic datasets to learn valid/invalid joint rotations in the swing and twist joint space and a discretized lookup-map for each joint is created. Similar to Akhter's approach, this method provides a discontinuous function. A comparison between this method and Akhter's method is presented in \cite{Murthy2019}. Based on their comparison, since the dataset used (i.e., Human3.6M dataset) is focused on human everyday routine activities, and does not include all possible human configurations, specifically on the boundaries, the model learnt is an under-approximation of the real RoM. Additionally, the axis-angle representation of swing angles make them different from joint angles defined in human biomechanics science.

In \cite{deng2022} authors proposed an algorithm to produce a human-like posture for a 7-DoF humanoid robotic arm. An OCSVM model trained on shoulder and elbow joint angles from human motion data was used to derive the boundary for human-like posture. In their model, shoulder flexion, shoulder abduction and elbow flexion motions are considered independent, and shoulder rotation is considered to be a function of them. A linear regression model was trained to predict the ideal shoulder rotation angle. The algorithm uses the redundancy of the robotic arm to improve the robot's posture as the secondary task. Unfortunately, in \cite{deng2022} the assumption of independent motions is inaccurate and the joint angle extraction algorithm, and the hyperparameter optimization for SVM model is not addressed. 

In this paper, we address these flaws by using an Euler angle parametrization so that joints motion components are compatible with known arm motion definitions in biomechanics literature, and comprehensible to human/physicians-in-the-loop. The discontinuity and Gimbal lock problems are tackled by computing the Euler angles from pre-processed quaternions exported from software as explained in section \ref{sec: angles}. Further, we propose a RoM dataset construction strategy that considers a more inclusive set of anatomical factors compared to prior works, and propose an automatic hyper-parameter tuning strategy for RoM model learning. 

%%%%%%%%%%%%%%%%%%%%%%%%%%%%%%%%%%%%%%%%%%%%%%%%%%%%%%%%%%%%%%%%%%%%%%%%%%%%%%%%

\begin{figure*} [tbp]
\begin{minipage}{0.348\textwidth}
    \centering
     \includegraphics[trim={1cm 0 1.5cm 0},clip,width=\linewidth]{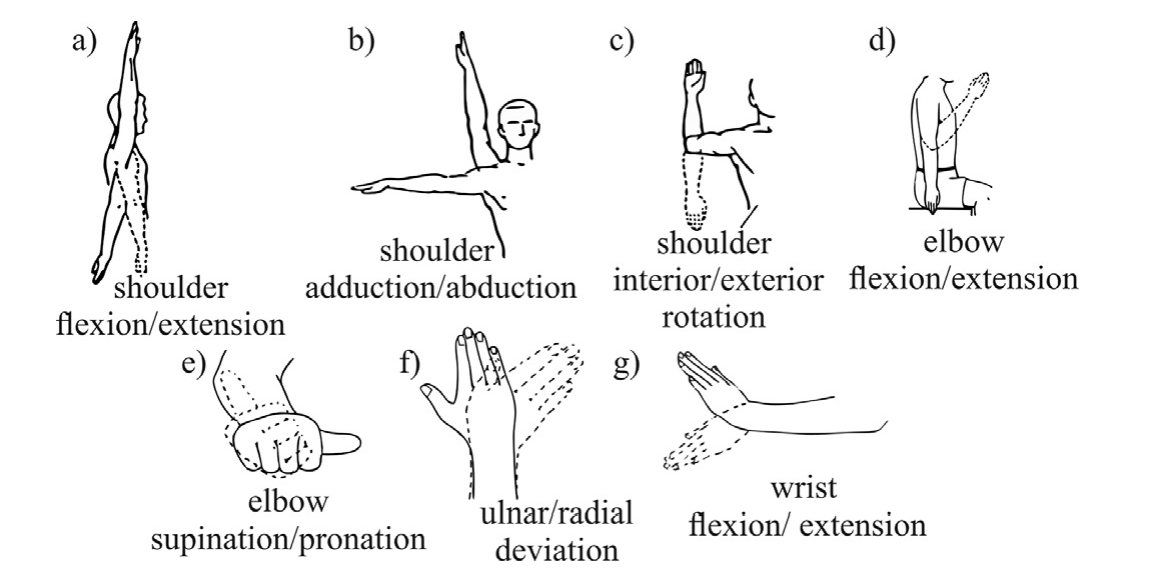} 
     \vspace{-20pt}
     \caption{RoM clinical evaluation guideline \cite{herbin2021}}
     \label{fig:clinical}
     % \vspace{-5pt}
\end{minipage}
\begin{minipage}{0.435\textwidth}
     \centering
     % \vspace{-5pt}
     \includegraphics[trim={0cm 0cm 7cm 1cm},clip,height=0.435\linewidth]{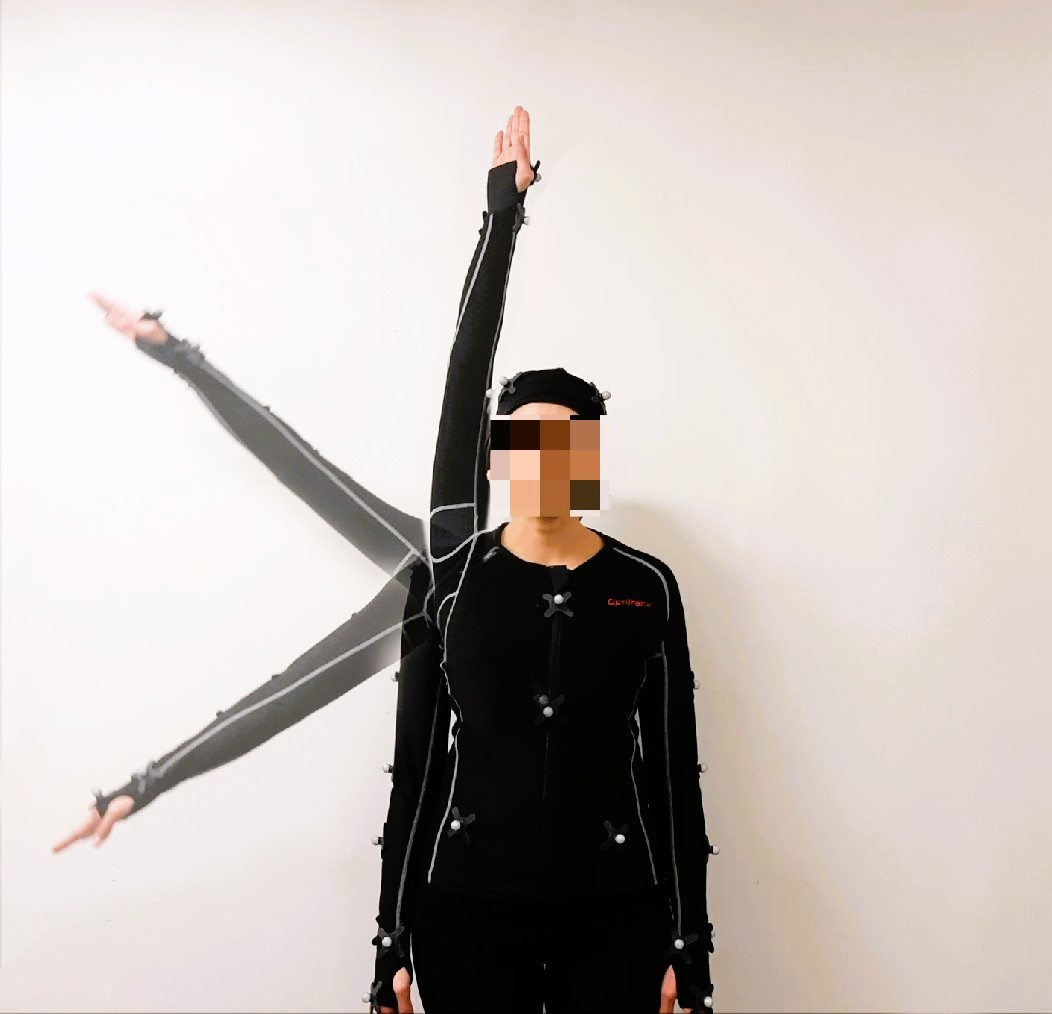}\includegraphics[trim={0cm 0cm 1cm 1cm},clip,height=0.435\linewidth]{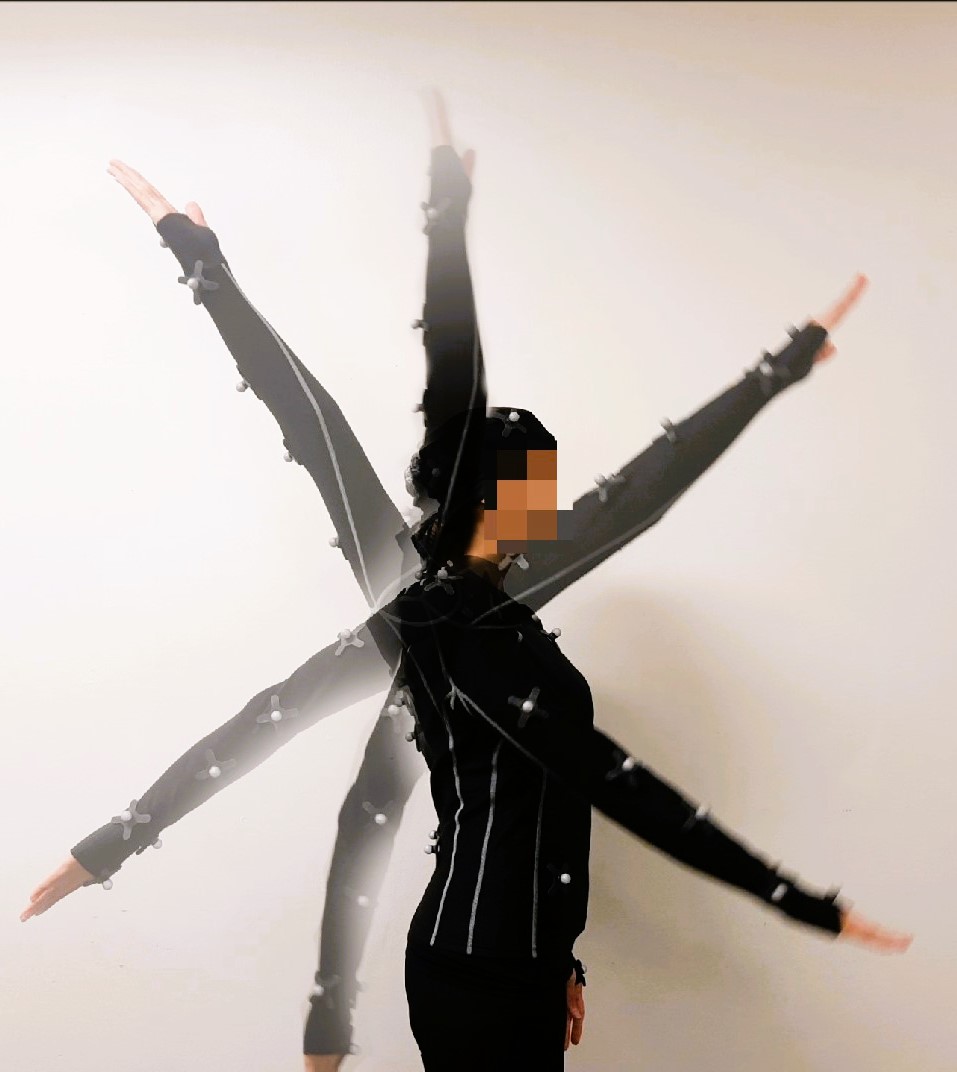}\includegraphics[trim={5cm 11cm 4.5cm 10cm},clip,height=0.435\linewidth]{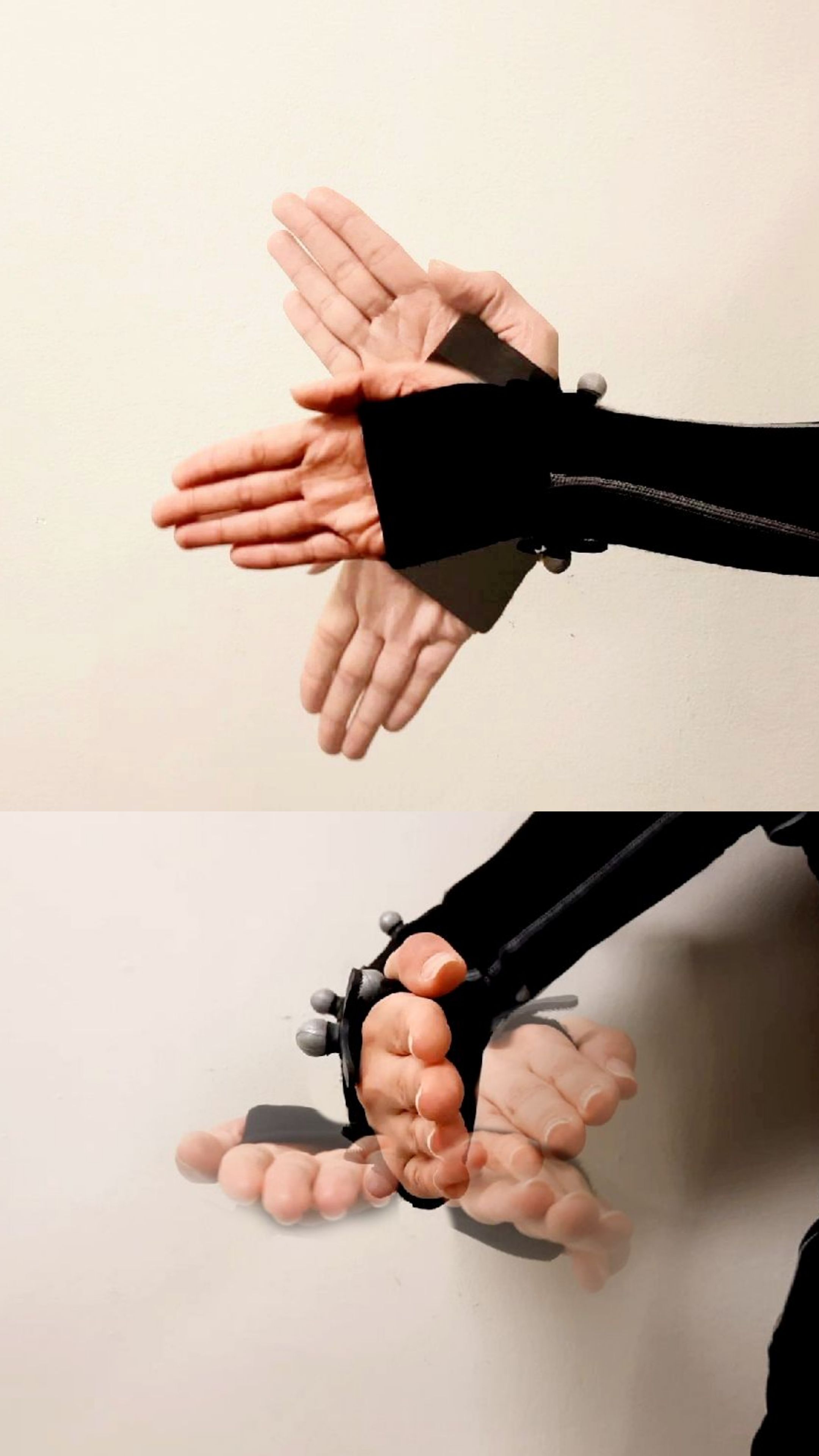}
     \vspace{-20pt}
     \caption{Subject performing shoulder abduction (left), shoulder extension (middle), wrist deviation and supination (right). \label{fig:subject-motions}}
\end{minipage}\hspace{5pt}\begin{minipage}{0.19\textwidth}
\centering
% \vspace{5pt}
\includegraphics[width=\linewidth]{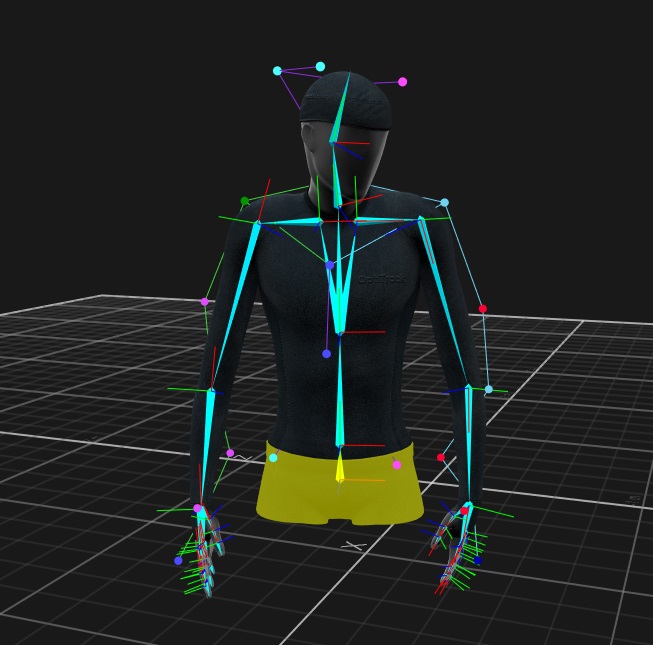}
% \vspace{-15pt}
\caption{Optitrack skeleton. \label{fig:OptiTrack}}
\end{minipage}
\vspace{-15pt}
\end{figure*}

\section{Learning Realistic RoM Boundary Functions}
\label{sec:learning-rom}
\subsection{RoM Dataset Construction}
\label{sec:data-construct}
To ensure that sufficient feasible configurations for the upper limbs are collected, reflecting constraints from all anatomical factors, the following experiments are designed. 
\subsubsection{Clinical RoM assessment} In this experiment, the clinical RoM evaluation \cite{dshsform} is implemented, in which each joint's RoM is explored by the subject independently (Fig.~\ref{fig:clinical} and \ref{fig:subject-motions}), and defined as $\mathcal{Q}_{\text{clinical}}= \{q^t\}_{t=1}^{M_{\text{{clinical}}}}$. The motions are demonstrated concurrently by an operator for subject's convenience. The motions are a) shoulder abduction/adduction, b) shoulder flexion/extension, c) shoulder horizontal abduction/adduction, d) shoulder rotation (with elbow $90^{\circ}$), e) elbow flexion/extension, f) elbow supination/pronation, g) wrist flexion/extension, and h) wrist deviation.

\subsubsection{Free Exploration} In this experiment each subject is asked to explore their upper limb RoM by discovering the space around them, while an operator demonstrate the motion to make sure all the possible configurations are collected, $\mathcal{Q}_{\text{exploration}}= \{q^t\}_{t=1}^{M_{\text{{exploration}}}}$ where $q^t\in\mathbb{R}^7$ is the $t$-th joint angle vector of the subject's upper limb and $M_{\text{{exploration}}}$ the number of samples in the free exploration dataset. 

The data collected from experiments 1 and 2 are combined as the RoM samples, $\mathcal{Q}_{\text{ROM}} = \{\mathcal{Q}_{\text{clinical}}, \mathcal{Q}_{\text{exploration}}\}$.

\subsection{Joint Angles Extraction} \label{sec: angles}
In this paper, upper-limb DoF are defined as in traditional biomechanics literature to be comprehensible for the human (e.g., physician, designer, etc.) in the loop. Joint angles are represented as Euler angles, and are extracted from motion capture data exported from Motive software \cite{motive}. Data includes position of each bone and the unit quaternions $(w,x,y,z)$ representing the orientation of bone frames, all defined in global coordinates. To address the discontinuities of the Euler representation observed in other datasets \cite{Akhter2015,cmu}, we calibrated the motion capture software with the same initial pose (i.e., T pose) for all trials. The software solves the quaternion values in real-time and chooses the quaternion set that matches the previous frame in Gimbal lock instances. The quaternions at each instance are then converted to rotation matrices, $R=\text{quat2R}(w,x,y,z)$ with respect to the world frame.

An articulated human skeleton model is a kinematic chain including bone segments connected by joints. Using the skeleton data, an upper-body kinematic chain including the hip, sternum, right and left upper-arms, right and left forearms, and right and left hands is built. We modeled each upper limb as a 7 DoF kinematic chain, with $\boldsymbol{q} =[q_1, ..., q_7]\in\mathbb{R}^7$ defined as shoulder abduction/adduction, shoulder flexion/extension, shoulder rotation (twist), elbow flexion/extension, elbow supination/pronation, wrist flexion/extension, and wrist deviation, respectively. The frames for shoulder, elbow, and wrist are set following the International Society of Biomechanics (ISB) recommendations \cite{wu2005isb}. 

To extract the relative angles between the upper limb joints, the following procedure is conducted. The relative rotation matrix between two joint frames, can be obtained using Eq.~\ref{eq:relR}, where $R_{prox}^{0}\in SO(3)$ is the rotation matrix of the proximal frame with respect to the global coordinates, and $R_{dist}^{0}\in SO(3)$ is the rotation matrix of the distal frame with respect to the global coordinates.
\begin{equation} \label{eq:relR}
R_{dist}^{prox} = {(R_{prox}^{0})} ^T   R_{dist}^{0}
\end{equation}
The joint angles $q_1, ..., q_7$ are extracted by converting the relative rotation matrices to Euler angles. The Euler sequence used is the `ZXY' sequence as recommended by the International Society of Biomechanics (ISB) \cite{wu2005isb}. The zero values of an upper limb DoFs is defined as demonstrated in Fig.~\ref{fig:OptiTrack} (right), when the upper limb is parallel to trunk and elbow is fully extended, with palm pointing towards the body.

\vspace{-1.5pt}
\subsection{Learning RoM Boundary Function}
\vspace{-1.5pt}
In this paper, one-class support vector machine (OCSVM) is employed to learn the RoM boundary function due to the availability of only one class of data (i.e., feasible joint angles), nonlinear characteristics of the data, and the goal to learn a differentiable boundary function. OCSVM is an unsupervised learning algorithm introduced by Schölkopf et al. \cite{scholkopf2000}, and it has been primarily used for anomaly detection. The OCSVM algorithm
finds a decision boundary that separates the data points from the origin. 
One advantage of an SVM algorithm is its ability to classify non-separable data using the kernel trick. This means that data points which cannot be linearly separated in their original space, are lifted to a high-dimensional feature space through a non-linear function $\phi$, where there can be a straight hyperplane that separates the data points from the origin. The hyperplane presenting the boundary decision in the feature space is defined linearly as $w^T\phi(q^i)-\rho=0$, with weight vector $w$ and offset term $\rho$. \begin{equation} \label{eq:decision1}
    f(q) = \text{sign}\Big(\big(w^T\phi(q)\big) - \rho \Big) 
\end{equation} The sign of the function can then be used to predict whether a new data point is inside the class ($+1$) or an outlier ($-1$).

To maximize the margin, subject to the constraint that all data points lie on the positive side of the hyperplane, i.e., $w^T\phi(q_i) - \rho \geq 0$, the following the convex optimization problem is solved,  
\begin{equation} \label{eq:ocsvm}
\begin{aligned}
& \underset{w,\xi,\rho}{\text{min}}
& & \frac{1}{2}\|w\|^2 + \frac{1}{\nu m}\sum_{i=1}^{m}\xi^i - \rho
\\
& \text{s.t.}
& &  \big(w^T\phi(q^i)\big) \geq \rho - \xi_i; ~~~ \xi_i \geq 0, \text{ for } i=1,\ldots,m,
\end{aligned}
\end{equation}
To prevent the SVM classifier from overfitting with noisy data, slack variables $\xi_i$ are introduced to allow some data points to lie within the margin. The hyperparameter $\nu$ is a value between 0 and 1, and determines the trade-off between maximizing the margin and the number of training data points within that margin, and controls the proportion of data points that are allowed to be misclassified.

Deriving the dual form of Eq. \ref{eq:ocsvm} and using the kernel trick $k(q,q^i) = \phi(q)^T\phi(q^i)$, the decision function in Eq.~\ref{eq:decision1} is transformed to the following support vector expansion:
\begin{equation} 
\label{eq:decision2}
   f(q) = \text{sign} \left( \sum_{i=1}^{m}\alpha_i k(q,q^i) - \rho\right)
\end{equation}
We formulate the RoM boundary function, $\Gamma(q):\mathbb{R}^N\rightarrow\mathbb{R}$ Eq. \ref{eq:gamma-definition}, as the continuous signed term of Eq.~\ref{eq:decision2}:
\begin{equation}  \label{eq:decision}
   \Gamma(q) = \left( \sum_{i=1}^{m}\alpha_i k(q,q^i) - \rho\right)
\end{equation}
with $q^i$ being support vectors selected from the training dataset $\mathcal{D}_{ROM}$ and $k(q^i,q^j)$ the Radial Basis function (RBF), 
$
   k(q^i, q^j) = \exp \left(-\frac{1}{2\sigma^2} ||q^i- q^j||^2 \right)
$
where $q^i$ and $q^j$ are two data points, $||q^i- q^j||$ is the Euclidean distance between them, and $\sigma$ is a hyperparameter, known as the RBF kernel scale, that controls the smoothness of the decision boundary.
% \vskip 0.5cm

\subsection{OCSVM Hyperparameter Tuning}
The main challenge in learning a boundary function for one-class classification, is the sensitivity of the model to hyperparameters, which are the kernel scale ($\sigma$) and $\nu$ in OCSVM. Inappropriate hyperparameters can result in overfitting and underfitting as shown in Fig.~\ref{fig:overunder}. These parameters influence the construction of a classification model and control how loosely or tightly the decision boundary encapsulates the training data. Hence, they need to be carefully selected based on the characteristics of the data and the desired behavior of the boundary. For instance, small $\sigma$, will generate a tight decision boundary that yields to a complex over-fitted classifier that is not generalizable. On the other hand, large $\sigma$ will result in underfitting since it generates a loose decision boundary that results in a model which classifies most of the new samples as positive class \cite{anaissi2018}. 
$\nu$ is an upper bound on the fraction of outliers and a lower bound on the fraction of support vectors (SV). A smaller $\nu$ will result in a bigger boundary, and similar to $\sigma$, $\nu$ also determines the trade-off between the covered region and possible overfitting problem \cite{zhuang2006}. There are numerous hyperparameter selection methods for binary-class SVM, yet, they do not apply to OCSVM due to the lack of negative instances \cite{xiao2014,anaissi2018}. Grid search has been widely used in SVM parameter selection and is regarded as the most reliable approach. In this paper, for each $q=[q_i, q_j]\in \mathbb{R}^2$ set, the RoM boundaries are learnt and tuned using OCSVM with constrained grid search to gain a smooth boundary that tightly encloses the data, while avoiding overfitting and underfitting. A sequential grid search is conducted. As the first step, the OCSVM is implemented on a wide range of hyperparameters $\nu\in \left(0,1\right]$, and $\sigma\in \left[10^{-3},10^{3}\right]$ to avoid further assumptions. Then the grid search is narrowed down by ruling out the unacceptable regions. This process is repeated until no considerable difference is observed in the learnt boundary by downsizing the grids. For the case of our data set and the inherent characteristics of the upper limb RoM, the following constraints are considered in assessing the SVM performance for each ($\nu, \sigma$) pair in the grid:

\begin{figure} [!tpb]
     \centering
     \includegraphics[width=0.48\textwidth]{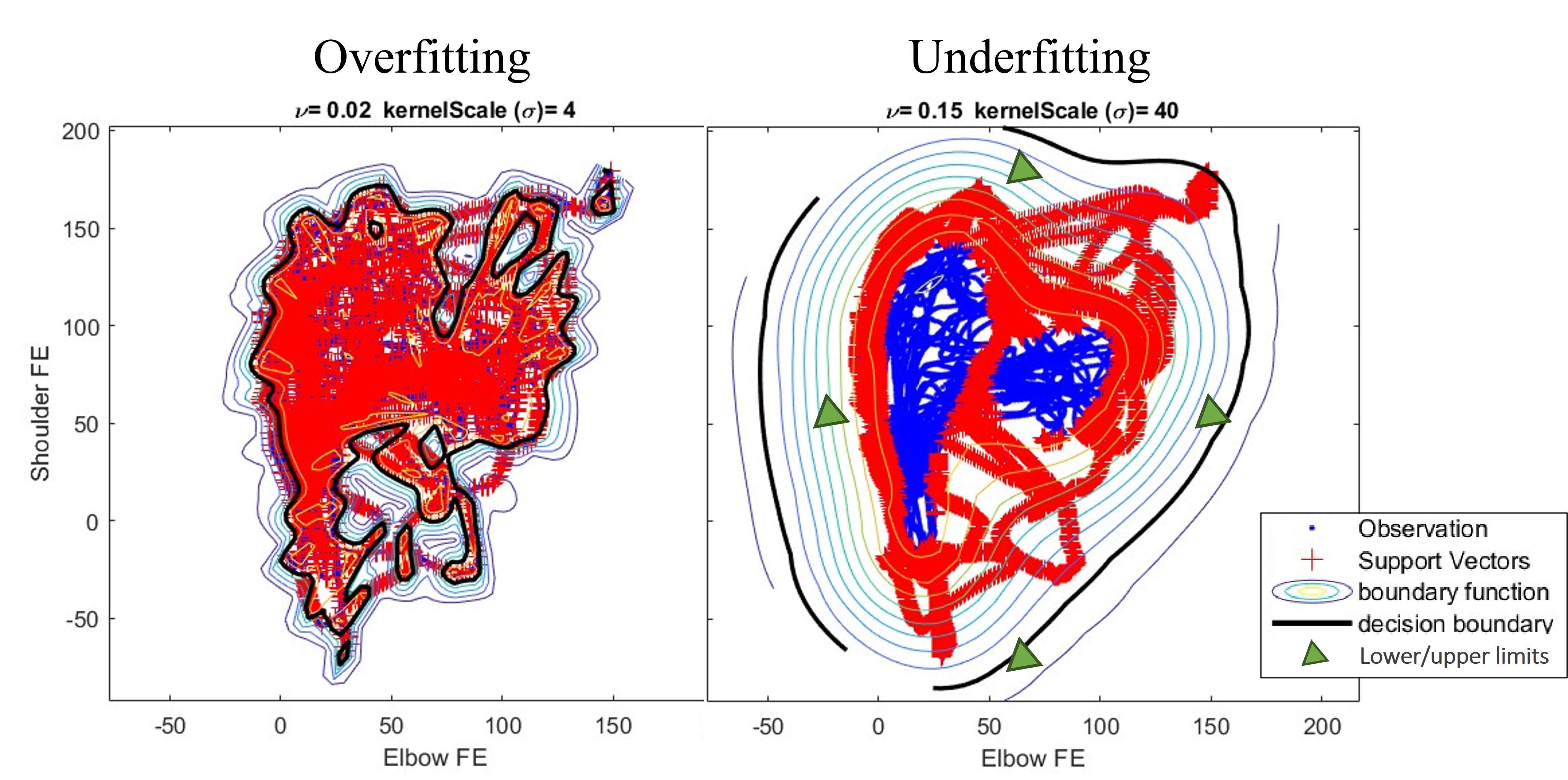} 
     \vspace{-15pt}
     \caption{Example of overfitting \textbf{(left)} and underfitting \textbf{(right)} RoM boundary learning with differing combinations of hyper-parameters $\nu$ and $\sigma$. Blue points represent training data collected as in Section \ref{sec:data-construct}, red cross ($+$) are support vectors, isolines depict level-sets of $\Gamma(q)$ with Eq. \ref{eq:decision}, bold black line represents the RoM boundary when $\Gamma(q)=0$ and the  green triangles ($\triangle$) are known upper/lower joint limits used to evaluate hyper-parameters.  \label{fig:overunder}}
     \vspace{-15pt}
\end{figure}

\textbf{1)} Additional random data is captured from the same participant, and is fed to the learnt model as test data. If $\Gamma(q)<0$ for any of the frames, the ($\nu, \sigma$) pair is excluded.

\textbf{2)} To avoid overfitting and assuring the smoothness of the boundary, we adopt the edge support vector (ESV) method introduced in \cite{anaissi2018}. According to \cite{anaissi2018}, a one-class SVM with a certain combination of $(\nu,\sigma)$ can be identified as an overfitted model when a large number of the SVs are located within the support of the sample distribution, as opposed to on the edge. In \cite{anaissi2018} for each SV a hard margin linear SVM is used to construct a hyperplane with its k-nearest neighbour samples, and if all the k-nearest neighbour samples lie on one side of this hyperplane, the SV is identified as an edge sample. If the number of interior SVs are more than a threshold, the model is not accepted. This threshold is added to accept some interior SVs when dealing with concave datasets. For the case of motion data captured from real human experiments, the k-nearest neighbours does not capture the points close to each SV due to the sparsity of the dataset in some regions. Hence, we modified the ESV method, so that instead of KNN search, neighbours are identified as the points inside a Euclidean ball of radius $r$ with the SV at the center, $\mathcal{B}_r(q^i) = \{q\in \mathbb{R}^N ~|~ l_2(q,q^i) \leq r\}$. We call this method modified edge support vector (M-ESV). The variables of M-ESV (i.e., $r$, number of misclassified neighbours, allowable number of inside SVs) is tuned for each ($q_i, q_j$) set based on the convexity and sparsity characteristics of the data.

\textbf{3)} To avoid underfitting and looseness of the boundary, similar to \cite{zhuang2006}, a number of known negative samples (majority class) are produced to validate the performance of the classifier. We produced points slightly passing the upper/lower joint limits of each dimension at the mean value of other dimensions (green triangles in Fig.~\ref{fig:overunder}). Since the captured data includes the minimum and maximum feasible value for each DoF, these instances should be classified as outliers. If any of the points are scored as a positive ($\Gamma(q)>0$), the ($\nu, \sigma$) pair is not accepted.   

\section{RoM-based Impairment Index (II) Metric}
\label{sec:impairment-metric}
For stroke patients with upper-limb disabilities, limited RoM is a common consequence due to muscle weakness, spasticity and joint contractures. As a result, it is expected that the volume enclosed by the RoM boundary be smaller in the impaired arm compared to their healthy arm, and the more impaired the arm, the more difference between the RoM volume. We introduce an objective Impairment Index (II) metric based on the ratio of impaired and healthy RoM volume to grade each patient's impairment as, \begin{equation} \label{eq:impindex}
II = \frac{V_{impaired}}{V_{healthy}}
\end{equation}
II metric can be used for diagnostic purposes, as well as tailoring the level of ``assistance'' given by the robot during robot-based therapeutic tasks. As the decision boundaries are learned, the RoM volumes, $V_{impaired}$ and $V_{healthy}$ can be calculated for the 7DoF RoM boundary. However, in order to derive a metric that is interpretable for the clinician in-the-loop and tunable based on the therapy exercises, we suggest the following formulation for volume derivation, 
\begin{equation} \label{eq:ROMvolume}
\begin{aligned}
\ V_{(impaired/healthy)} = \frac{1}{2} \ \sum_{i,j=1; i\neq j}^{7} \ C_{ij} \ V_{(im/he),ij}
\end{aligned}
\end{equation}
where $V_{im,ij}$ and $V_{he,ij}$ are the area of the ${ij}^{th}$ decision boundary for impaired and healthy arm respectively, and $C_{ij}$ are the weights corresponding to each 2 DoF pairs. This formulation provides the clinician with maneuverability to weigh the metric based on importance of DoFs. For instance, for therapeutic exercises that focus on the reachable workspace of a patient's arm, the DoFs corresponding to shoulder motions and elbow flexion need to be considered.
\vspace{-2.5pt}
\section{Experimental Results}
\label{sec:experiments}
\vspace{-2.5pt}
\subsection{Motion Capture Data Acquisition}
The data collection is implemented using 24 OptiTrack cameras and Motive Optical motion capture software \cite{optitrack}. The motion capture collection frequency is set to 100 Hz which is adequate for experiments involving human motion, since it provides accuracy while avoiding excessive amounts of data and high-frequency noise \cite{kruk2018, aylward2006}. In the Motive software \cite{optitrack}, \textit{skeleton assets} are used to create human skeletal models and tracking motions. Skeleton assets require additional calculations to identify and label markers to solve the skeleton model of the human. To that end, Motive uses pre-defined Skeleton Marker set templates, i.e., a collection of marker labels and their specific positions on a subject. 

For this study the Optitrack ``Conventional Upper'' skeleton marker set with 27 markers is used. This marker set template creates a skeletal model including 13 upper body bones (i.e. hip, abdomen, chest, right and left clavicles, right and left upper-arms, right and left forearms, and right and left hands), as well as three knuckle bones for each finger \cite{optitrack}. The finger joints are ignored in this paper. The solved upper-body skeleton model is shown in Fig.~\ref{fig:OptiTrack}.  

\subsection{Data Collection Experiments}
\begin{figure}[tbp]
     \centering
     \includegraphics[width=0.475\linewidth]{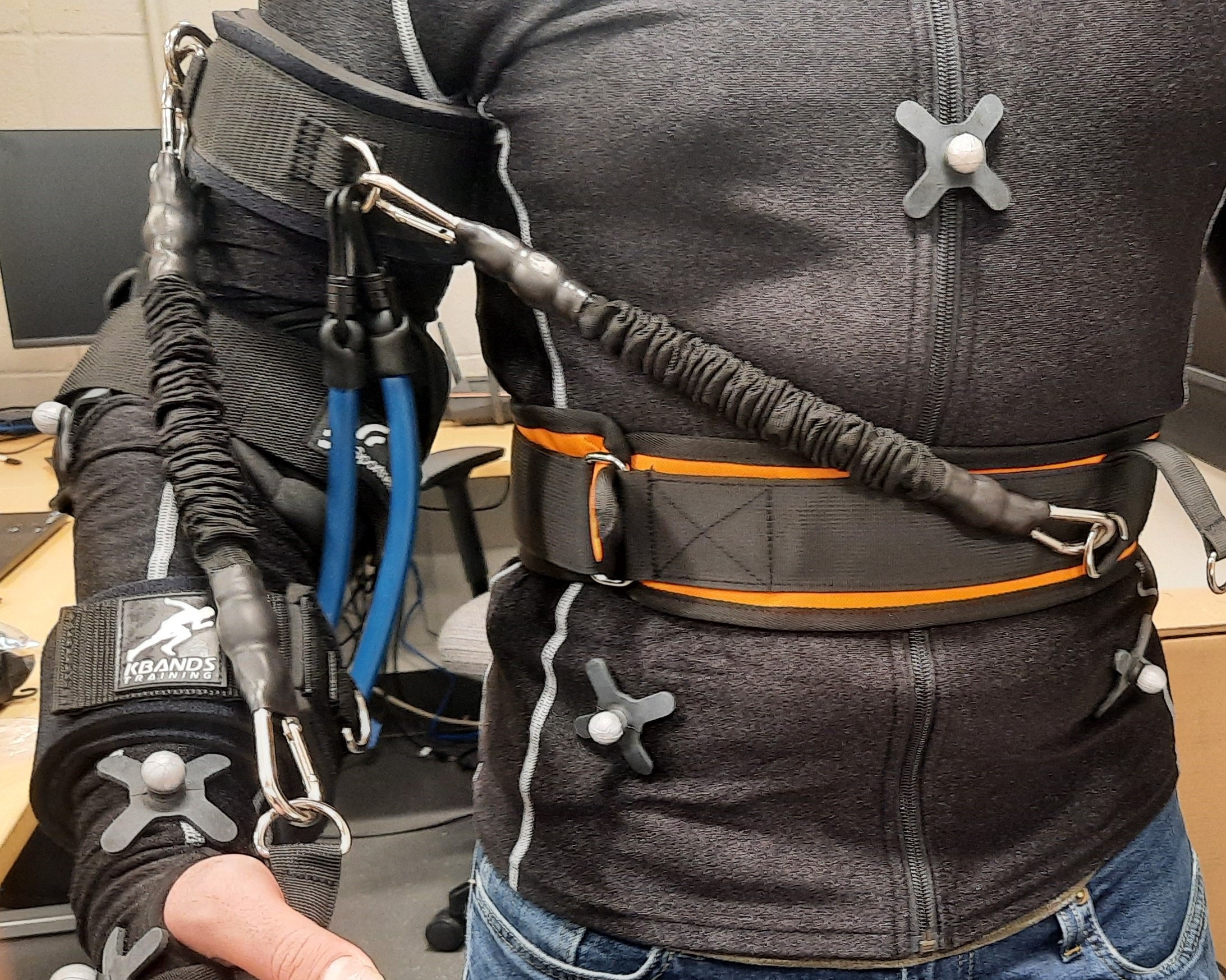} 
     \includegraphics[width=0.360\linewidth]{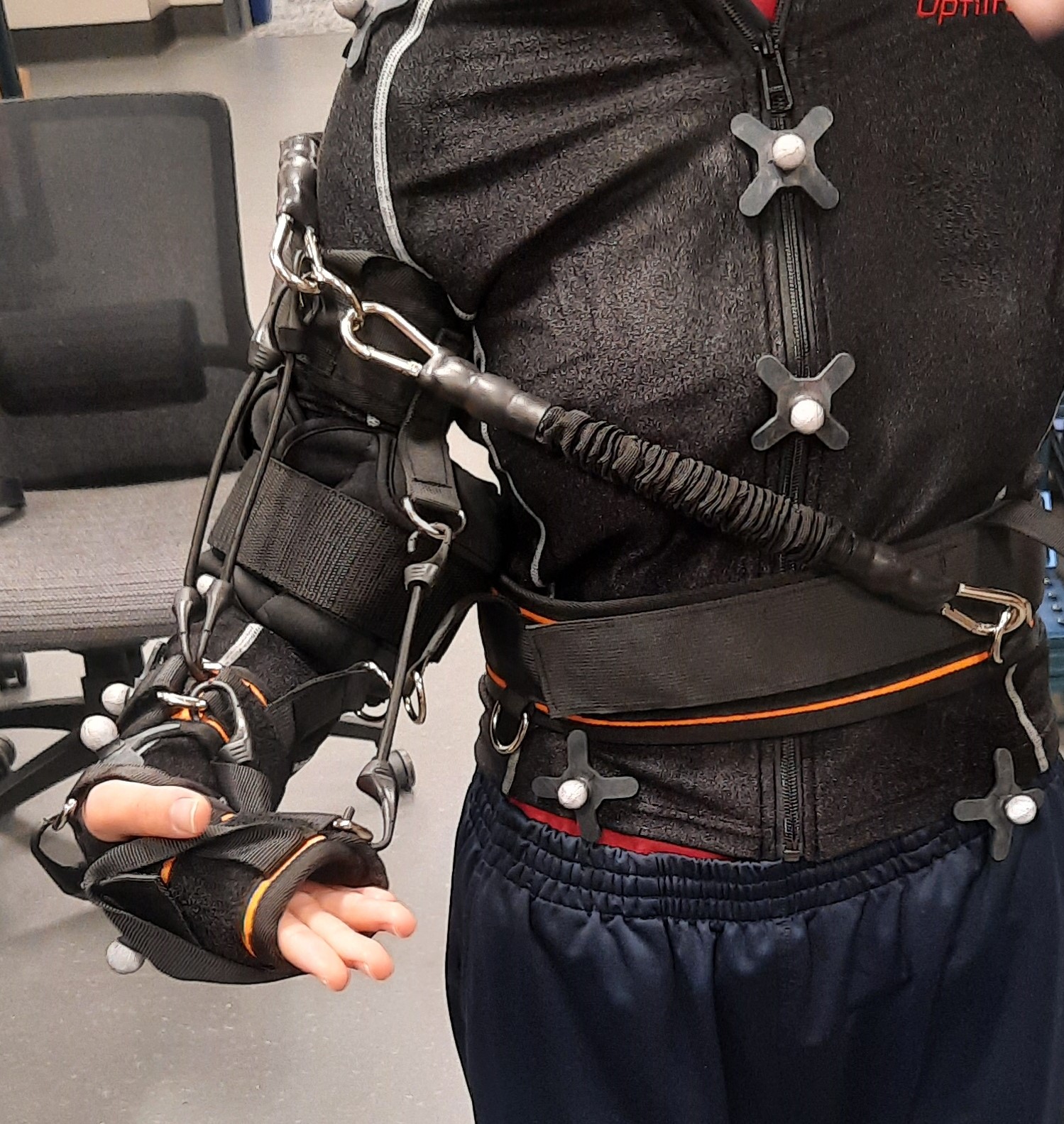}
      \vspace{-5pt}
     \caption{Weights and resistance bands added to emulate the disability.\label{fig:subs}}
      \vspace{-5pt}
\end{figure}

In stroke patients with hemiplegia one side of the body is impaired, leading to different RoM in the impaired and healthy arms. As described in Section \ref{sec:data-construct}, we propose to collect RoM data of each patient's arm. To validate the method proposed in this paper, no stroke patients are involved at this stage of the study. Instead, four healthy subjects participated in data collection and emulation of hemiplegia impairment and limited RoM caused by it. For each subject, extra weights and resistance bands were attached to participants' right arm as shown in Fig.~\ref{fig:subs} to emulate different levels of disability. Demographic and impairment information of participants is reported in Tab.~\ref{tab:subjects}. The number of healthy arm samples per participant is provided in Tab.~\ref{tab:PP}.

\subsection{Experiment Analysis}
The 7DoF human arm model can be kinematically decoupled into two sets, a set including $q_1, q_2, q_3,$ and $q_4$ responsible for the shoulder, elbow and wrist joints positions, and the set including $q_5, q_6, q_7$ responsible for the orientation of the hand. The motion of these two sets is either considered independent, or the dependency is not addressed in the literature. Additionally, in the experiments designed in this paper, not enough data is captured to analyze the dependency between the DoFs of these two sets. To our best knowledge, the data required for this analysis is not available in other datasets as well. Hence, we analyse the RoM data and boundary for shoulder ($q_1, q_2, q_3$) and elbow ($q_4$) joints. Considering that the RoM of ($q_1, ..., q_4$) suffice for posture correction in graphics, robotics and biomechanics applications, and in the majority of physical and occupational therapy sessions, we use the 4-dimensional RoM boundary function learned as $\Gamma(q_1, ..., q_4)$ from $\mathcal{D}_{RoM}$. 

\begin{table}[tbp]
  \centering
  \caption{Demographic information of participants}
  \vspace{-7pt}
  \label{tab:subjects}
  \begin{tabular}{|c|l|l|c|c|c|} 
    \hline
    & Simulated impairment
 & Sex & Age & Weight & Height \\
    \hline
    \#1 & severe   & M & 31 & 85 kg & 175 cm \\
    \hline
    \#2 & moderate & F & 25 & 50 kg & 160 cm \\
    \hline
    \#3 & moderate & M & 27 & 68 kg & 178 cm \\
    \hline
    \#4 & mild     & M & 27 & 70 kg & 173 cm \\
    \hline
  \end{tabular}
  \vspace{-15pt}
\end{table}

\begin{figure} [thpb]
     \centering
     \includegraphics[width=0.48\textwidth]{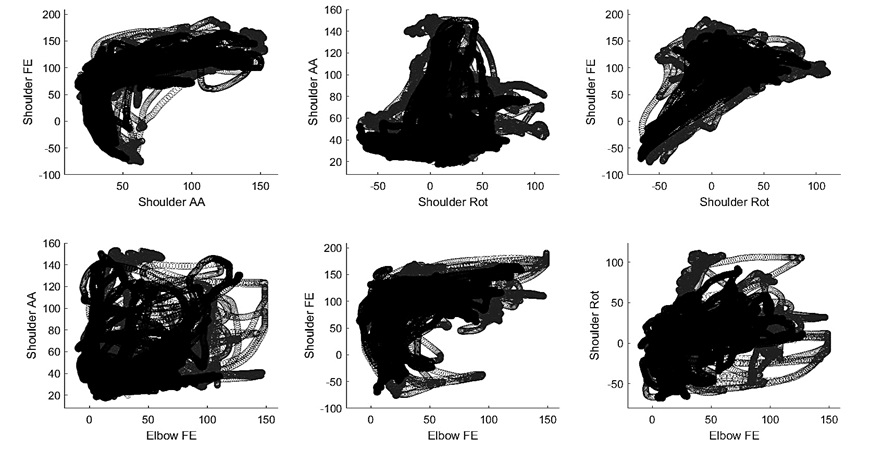} 
     \vspace{-15Pt}
     \caption{Visualization of DoF pairs: shoulder abduction, shoulder flexion, shoulder rotation, and elbow flexion (in degree). The inter- and intra-joint dependency of the arm's DoFs can be seen in these pairwise RoM data.}
     \label{fig:angs}
\end{figure}

For a query arm configuration, $\Gamma(q)$ can be used for human-like posture verification, and $\nabla\Gamma(q)$ can be used for realistic smooth human motion simulation, and control policy in HRI \cite{mirrazavi2018}. Fig.~\ref{fig:angs} shows the RoM data presented as the projection of 2d pairs of DoFs of shoulder abduction, flexion, and rotation, and elbow flexion for a subject's healthy arm. 

\begin{figure*} [thpb]
\begin{minipage}{0.375\textwidth}
         \centering
     \includegraphics[width=0.95\textwidth]{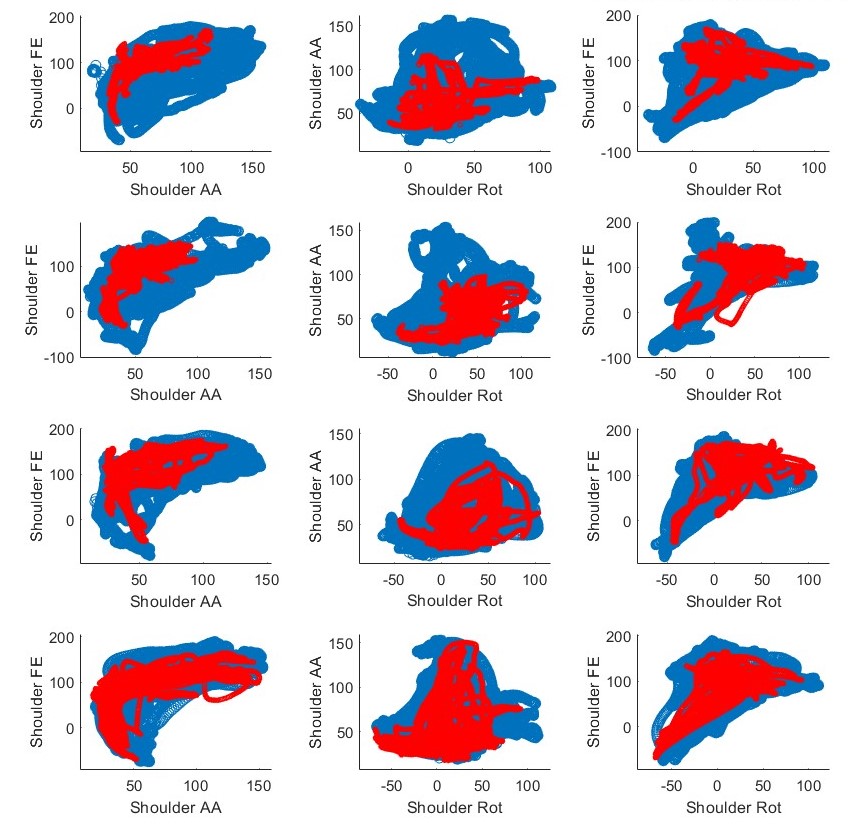} 
     % \caption{}
     % \label{fig:romImpairment}
\end{minipage}\begin{minipage}{0.62\textwidth}
         \centering
         \vspace{-2pt}
     \includegraphics[width=0.3\textwidth]{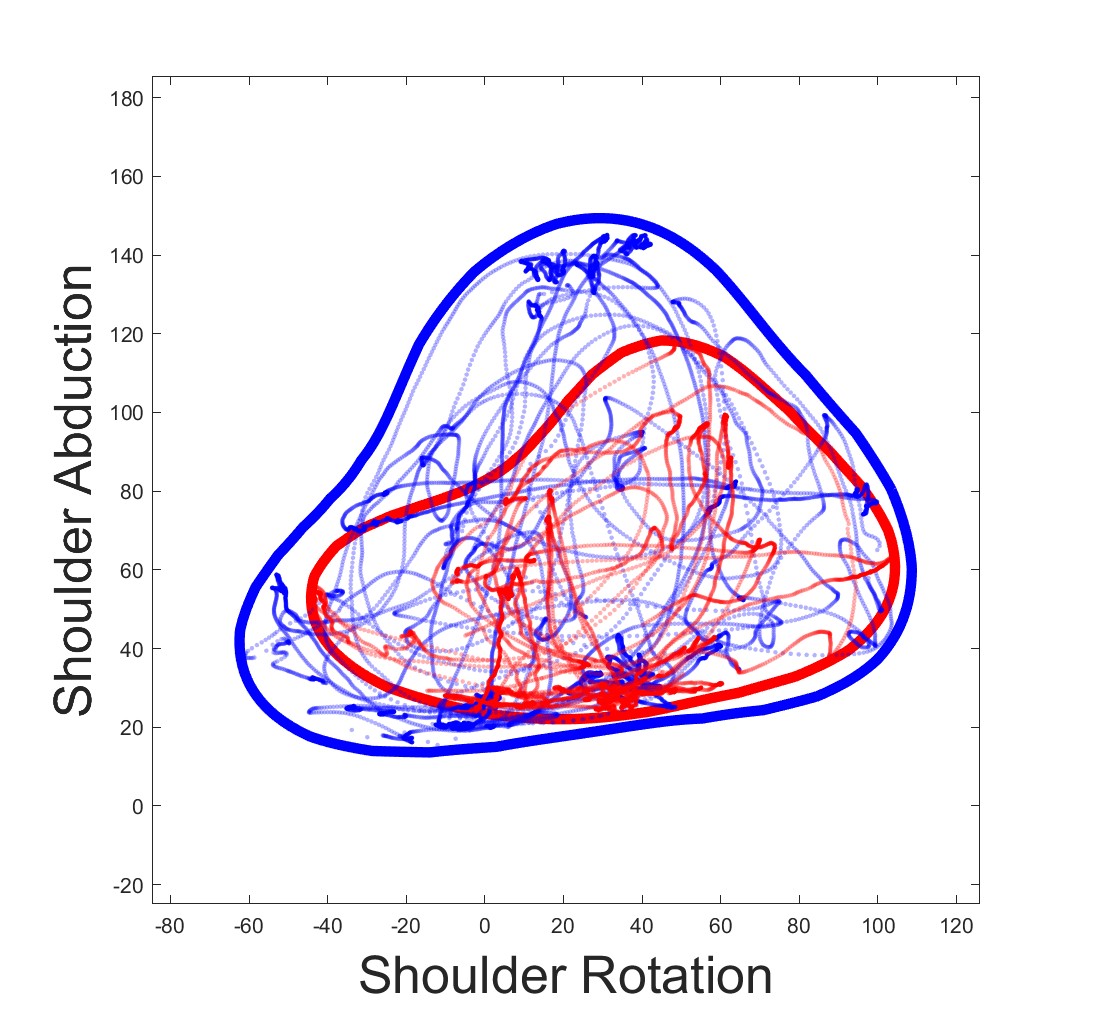} 
     \includegraphics[width=0.3\textwidth]{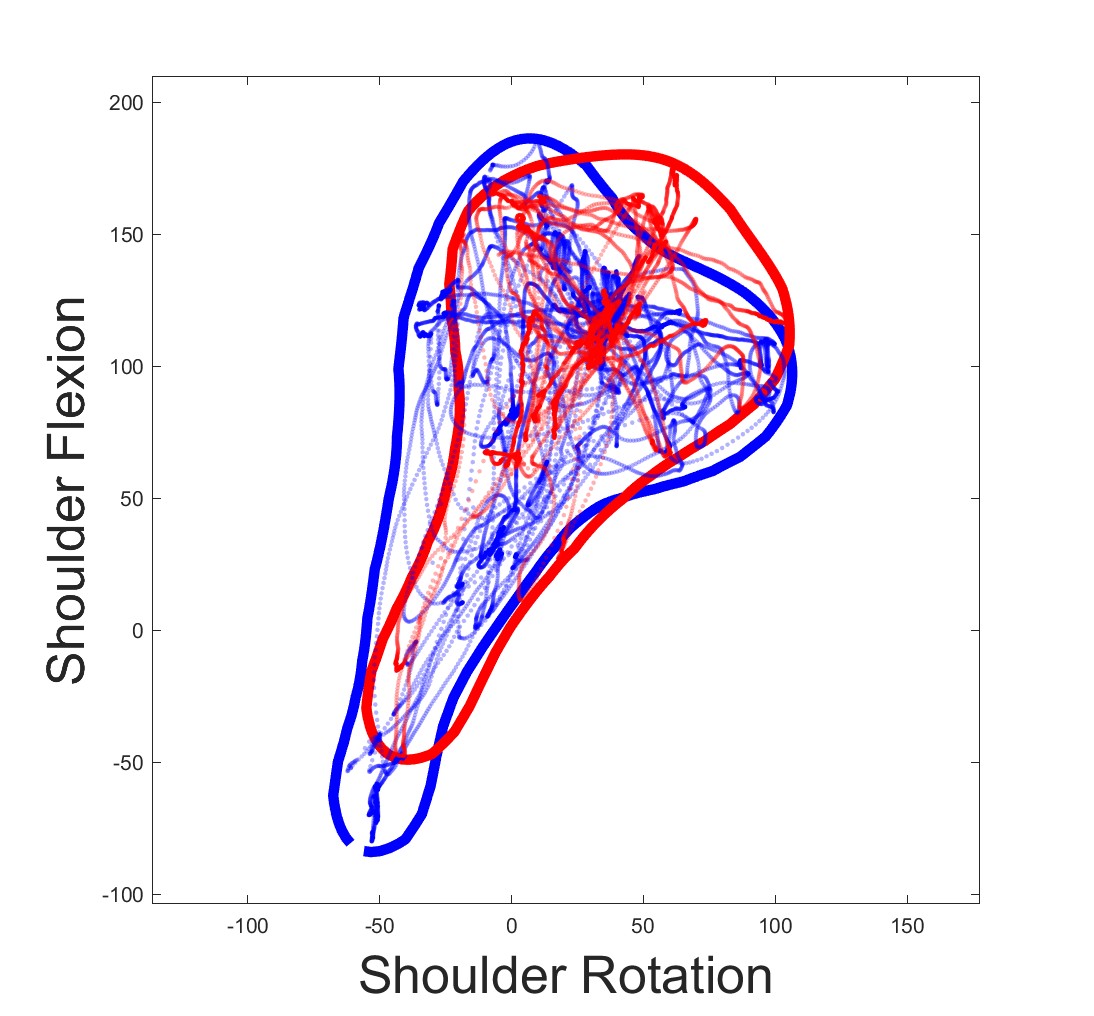} 
     \includegraphics[width=0.3\textwidth]{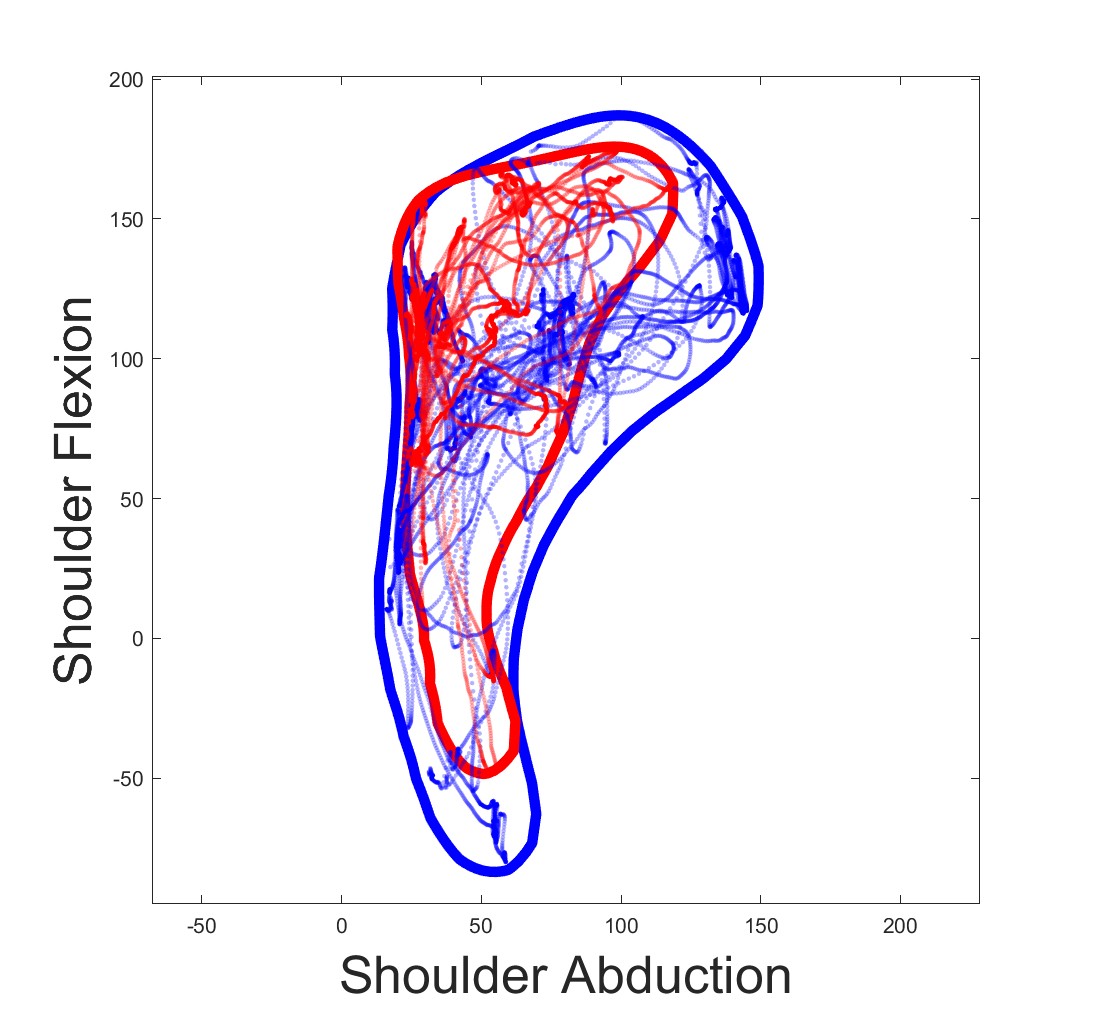} 
     \includegraphics[width=0.3\textwidth]{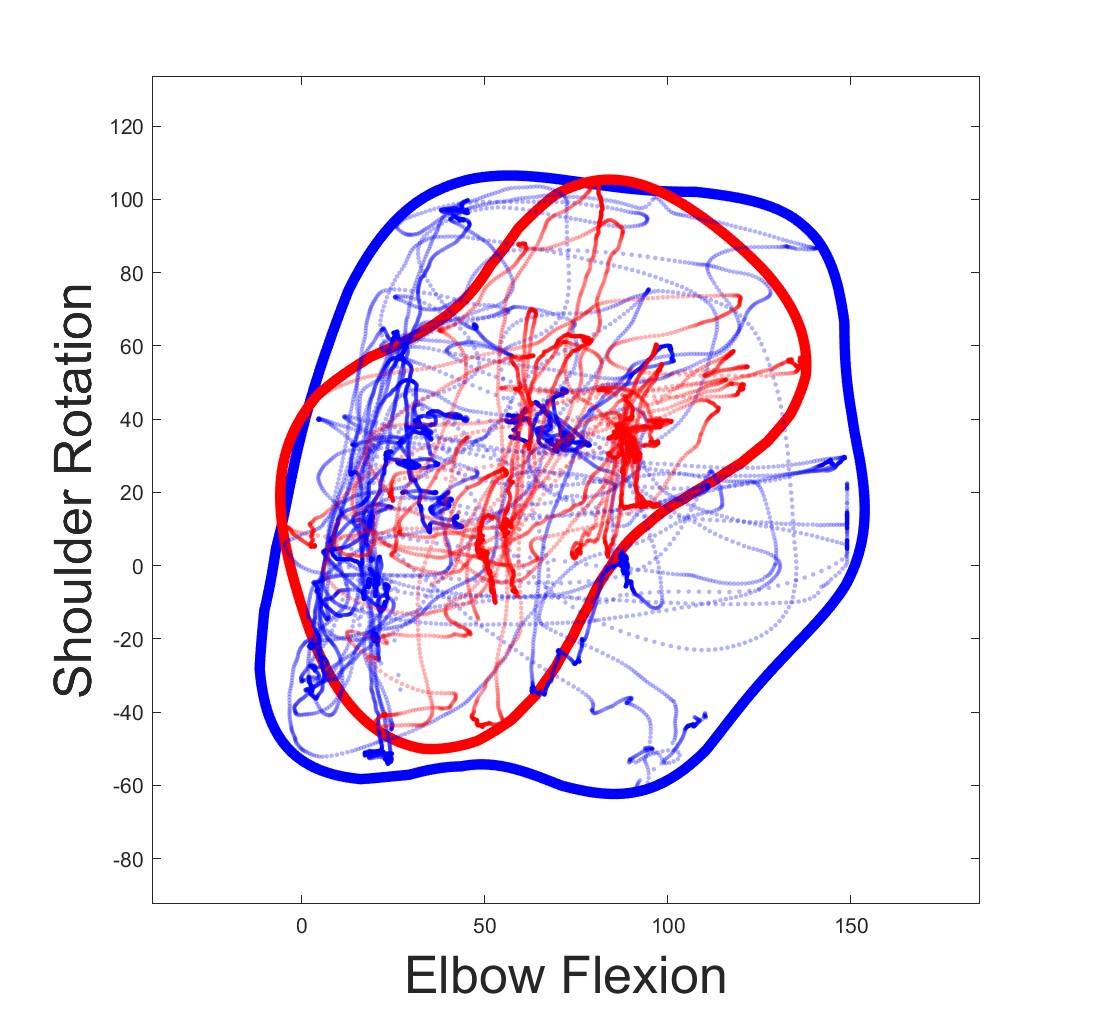} 
     \includegraphics[width=0.3\textwidth]{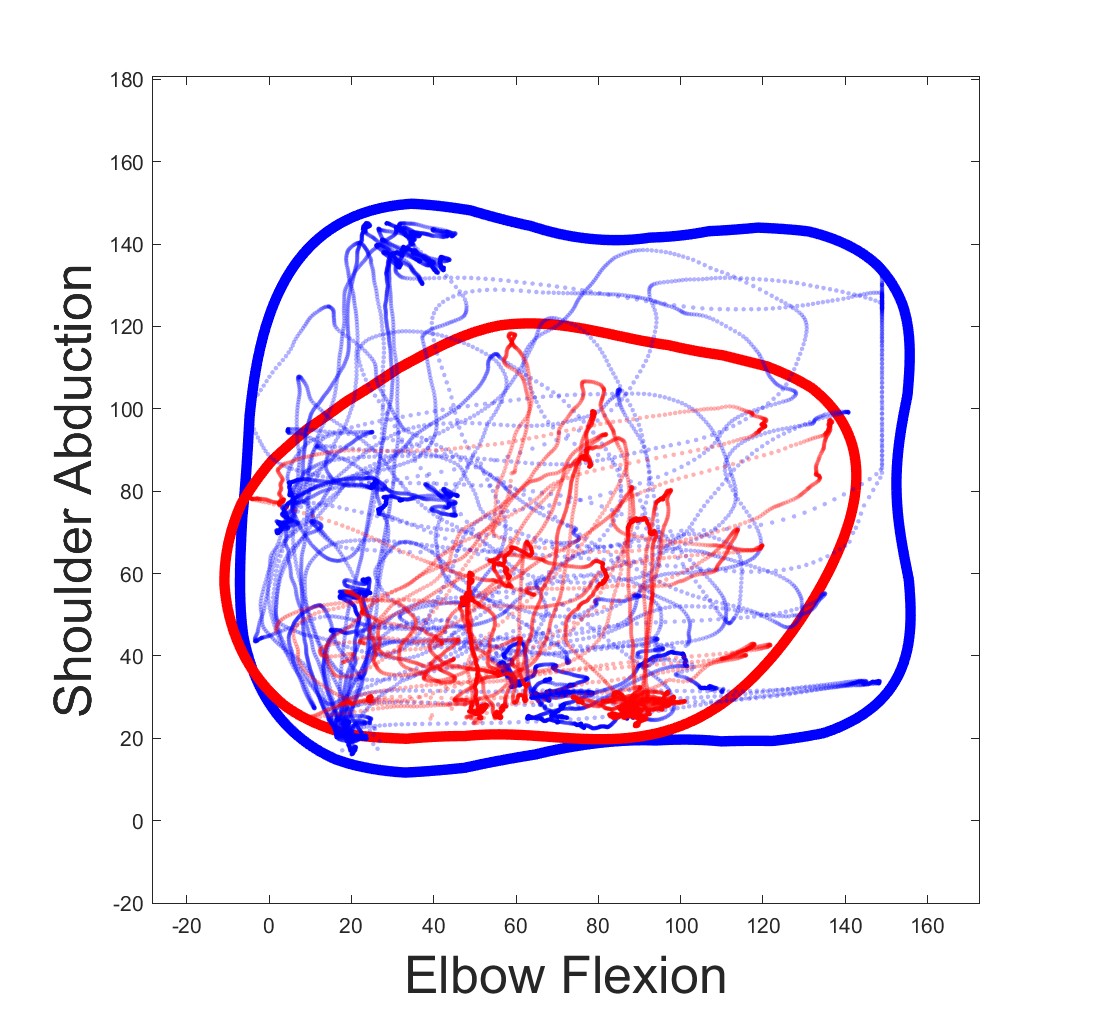} 
     \includegraphics[width=0.3\textwidth]{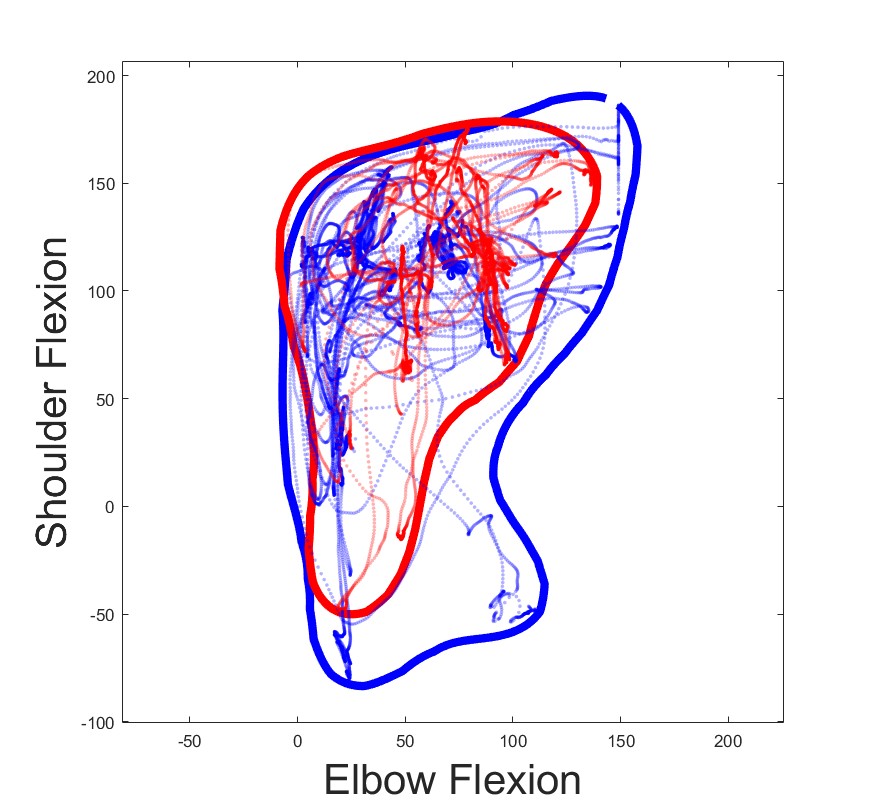} 
\end{minipage}
\caption{\textbf{(left)} Shoulder RoM data in different levels of disability for healthy (blue) and impaired (red) arms for individuals with severe (top) to mild (bottom) levels of impairment. \textbf{(right)} RoM data and learned boundaries for healthy (blue) and impaired (red) arms of subject 3 with moderate impairment. \label{fig:boundaries}}
\vspace{-15pt}
\end{figure*}

Fig.~\ref{fig:boundaries} (left) shows the influence of the level of impairment on the RoM dataset by comparing all the captured data of simulated impaired and healthy arms of each subject. The first to fourth rows of Fig.~\ref{fig:boundaries} (left) correspond to severe to mild levels of impairment respectively.

\subsection{RoM Boundary Learning Evaluation}
The hyperparameter tuning procedure proposed in this paper provides the acceptable ranges for ($\nu, \sigma$) pair upon those the OCSVM model can provide a smooth and inclusive boundary for each $q_i, q_j$ set. From the resulting acceptable ($\nu, \sigma$) pairs for each test, we can subjectively choose a $(\nu, \sigma)$ pair which results in a smooth and well-fitted boundary. Finally, the hyperparameter pair $(\nu=0.0075, \sigma=40)$ is chosen since it resulted in appropriate boundary for all $(q_i, q_j)$ sets, and for healthy and simulated impaired arm of all participants; with an average of $0.8299\%$ of data points as support vectors. The consistency in parameters may indicate that similar hyper-parameters might be shared by similar demographics. This can lead to improved hyper-parameter tuning and meta-learning strategies as more RoM datasets are collected. The resulting RoM boundaries of healthy and impaired arms for subject 3 with moderate impairment is shown in Fig.~\ref{fig:boundaries}. As it can be seen in Fig.~\ref{fig:boundaries}, the RoM boundaries for impaired arm (red) is limited with respect to the RoM boundaries for healthy arm (blue). These DoF sets are informative as they provide a visual description of training targets for the upper limb. The x-sectional volume difference is large for Shoulder Abd/Add versus Elbow FE suggesting this movement combination as a possible target for exercises or assistance during robot-assisted therapy. However, exceptions where RoM of impaired arm exceed the RoM of healthy may exist, e.g., in Fig.~\ref{fig:boundaries} (right) top-center. Every individual has different ROM in their right and left limbs due to different levels of flexibility, which can be caused by handedness, injury history, etc. This participant is right-handed, and their right arm was emulated impaired. As a result, the ROM of their dominant impaired arm exceeds the ROM of their non-dominant healthy arm, in some joint motions. Hence, hand-dominance needs to be considered when comparing the RoM boundaries and using the II metric.

\begin{table}[thpb]
 \vspace{-5pt}
  \centering
  \caption{Impairment Index for different levels of disability}\label{tab:ii}
  \vspace{-5pt}
  \begin{tabular}{|c|c|c|c|c|c|}
    \hline
    Participant & Disability & $V_{healthy}$ & $V_{impaired}$ & II \\
    \hline
    \#\ 1 & severe    & 12850.0 & 6229.6 & 0.4848 \\
    \hline
    \#\ 2 & moderate  & 13450.0 & 8621.4 & 0.6410 \\
    \hline
    \#\ 3 & moderate  & 12825.3 & 8380.7 & 0.6534 \\
    \hline
    \#\ 4 & mild      & 13167.9 & 9328.4 & 0.7084 \\
    \hline
  \end{tabular}
   \vspace{-10pt}
\end{table}

\subsection{Impairment Index}
To measure the level of impairment we calculate the Impairment Index (II) for each participant using Eq.~\ref{eq:impindex} and Eq.~\ref{eq:ROMvolume} with $C_i^j=1$ between healthy and impaired arms.
As seen in Table.~\ref{tab:ii}, the RoM boundary volume itself does not indicate the impairment level as each individual has different RoM due to different level of flexibility. The proposed II metric clearly indicates the simulated impairment level of the subjects. As expected, the impairment index decreases proportional to the severity of the impairment increases. 

\subsection{Comparison with state-of-the-art methods}
\begin{figure*}[!tpb]
\begin{minipage}{0.495\textwidth}
     \centering
     \includegraphics[trim={0.65cm 0cm 13.5cm 0.5},clip,width=0.85\linewidth]{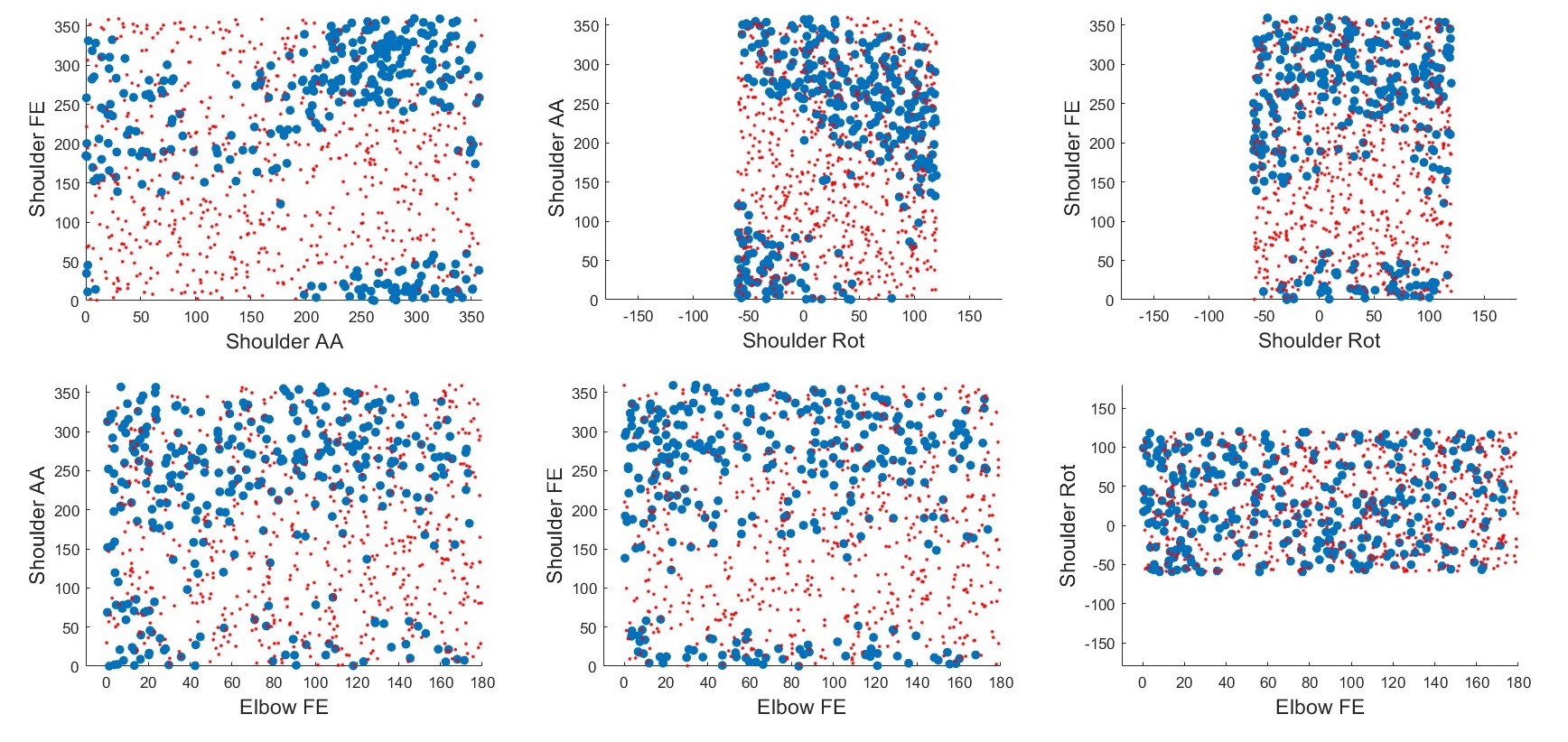} 
     \vspace{-15pt}
     \caption{Non-contiguous clustering in Jiang's method \cite{Jiang2018} due to Euler angle parametrization. random data classified as (blue/red) for valid/invalid. \label{fig:Jiang}}
     \vspace{-10pt}
\end{minipage}\hspace{2pt}
\begin{minipage}{0.495\textwidth}
     \centering
     \includegraphics[trim={0cm 0.5cm 20cm 0},clip,width=0.85\linewidth]{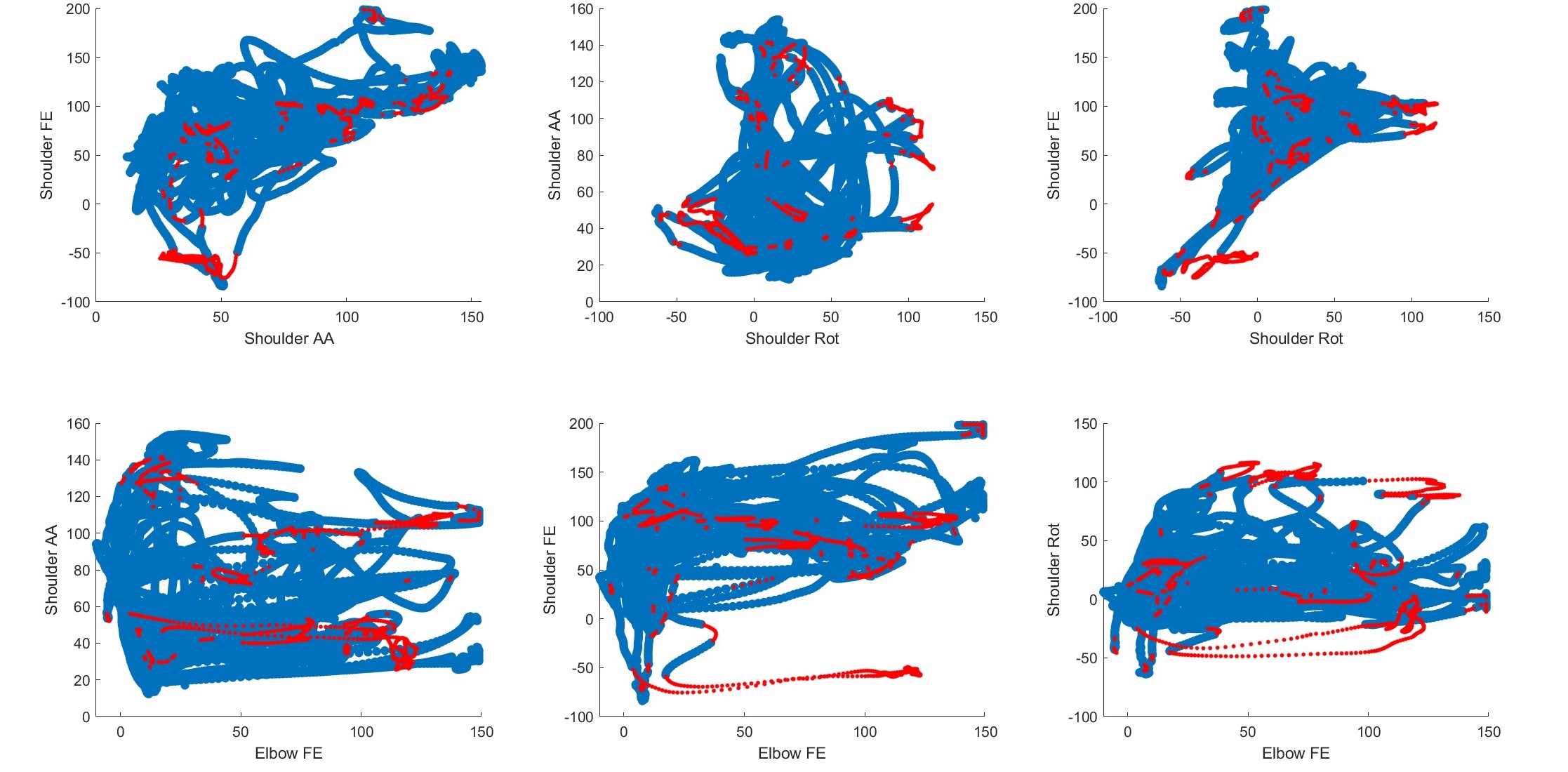} 
     \vspace{-10pt}
     \caption{RoM data of subject 2, healthy arm, classified by PosePrior \cite{Akhter2015} prior. observations classified  (blue/red) for valid/invalid   \label{fig:PPvsROM}}
     \vspace{-10pt}
     \end{minipage}
     \vspace{-5pt}
\end{figure*}

Jiang \cite{Jiang2018} and Ahkter's \cite{Akhter2015} methods are the only approaches that can solve a similar problem to ours. Nevertheless, a direct comparison on our datasets is not straightforward as they use different joint angle parametrizations. The joint angle modality of `PosePrior' dataset lacks consistent calibration across trials, hence, exhibits varying joint angle values for the same human posture, which led to discontinuities within the dataset, as also addressed in \cite{Murthy2019}. Further, Jiang's method which builds upon PosePrior, induces non-contiguous clusters due to the Euler angle parametrization, as shown in Fig.~\ref{fig:Jiang}. However, since the goal of PosePrior \cite{Akhter2015} is to provide a standard prior learned to classify the validity of a human-like pose, we were able to classify our captured data (i.e., exported joint positions) employing PosePrior's prior. The resulting classified data for subject 1 is shown in Fig.~\ref{fig:PPvsROM}. As shown, some of the captured observations are classified as invalid by PosePrior. For each subject, the ratio of the misclassified (i.e. False Negative) points to the total points is reported as miss-classification rate in Table.~\ref{tab:PP}. From the misclassified regions, it can be concluded that the model proposed in this study is more inclusive, especially around extreme shoulder angles and non-common configurations such as the “scratching the back” pose. 

\begin{table}[thpb]
 % \vspace{-3pt}
  \centering
  \caption{Mis-classification rate of RoM data using PosePrior prior}\label{tab:PP}
   \vspace{-5pt}
  \begin{tabular}{|c|c|c|c|c|}
    \hline
    Participant & 1 & 2 & 3 & 4 \\
    \hline
    Number of samples & 88523 & 23496 & 19730 & 15910 \\
    \hline
    Misclassification rate & 36.9  \% & 8.3 \% & 6.6  \% & 5.7  \% \\
    \hline
  \end{tabular}
   \vspace{-10pt}
\end{table}

%%%%%%%%%%%%%%%%%%%%%%%%%%%%%%%%%%%%%%%%%%%%%%%%%%%%%%%%%%%%%%%%%%%%%%%%%%%%%%%%
\section{Conclusions \& Future Work}
\vspace{-3pt}
In this paper, a realistic human arm RoM is learned from motion capture data considering intra-joint and inter-joint dependencies. The RoM boundary is represented as a continuously differentiable function $\Gamma(q)$ which returns a positive value inside the boundary (i.e., feasible configurations), a negative value outside the boundary, and zero on the RoM boundary. This function is learnt by OCSVM algorithm with an efficient hyperparameter tuning strategy. We learn RoM boundaries of healthy and impaired arms of participants. The volumes enclosed by the RoM boundaries define a novel objective metric to quantify the impairment level of stroke patients. The proposed algorithm is generic and can be utilized for learning the realistic RoM for other joints of any high-dimensional human model. Besides impairment diagnosis, we envision the learnt function $\Gamma(q)$ to be used in real time for (1) determining human-like joint configuration, (2) determining the feasibility of a task for an individual, and (3) enforcing the anatomic constraints in solving the inverse kinematics of the human motion. Due to the continuous differentiable kernel used in our OCSVM formulation, we can obtain the gradient of the boundary function, $\nabla\Gamma(q)$, which can be used to provide smooth and meaningful motion as the joints reach their limits, using constraint forces in simulations and in HRI control policy considering the human capabilities. Improving model complexity and learning efficient is our next step to move towards the aforementioned applications.

%%%%%%%%%%%%%%%%%%%%%%%%%%%%%%%%%%%%%%%%%%%%%%%%%%%%%%%%%%%%%%%%%%%%%%%%%%%%%%%%

\bibliographystyle{IEEEtran}
\bibliography{rom.bib}
\end{document}